\documentclass{article}

\usepackage[preprint]{neurips_2024}

\usepackage[utf8]{inputenc} % allow utf-8 input
\usepackage[T1]{fontenc}    % use 8-bit T1 fonts
\usepackage{hyperref}       % hyperlinks
\usepackage{url}            % simple URL typesetting
\usepackage{booktabs}       % professional-quality tables
\usepackage{amsfonts}       % blackboard math symbols
\usepackage{nicefrac}       % compact symbols for 1/2, etc.
\usepackage{microtype}      % microtypography
\usepackage{xcolor}         % colors
\usepackage{amssymb}
\usepackage{amsmath}
\usepackage{multirow}
\usepackage{threeparttable}
\usepackage{graphicx}

\title{Curate-Train-Refine: A Closed-Loop Agentic Framework for Zero Shot Classification}

\author{%
  Gaurav Maheshwari \\
  Diabolocom\\
  \texttt{gaurav.maheshwari@diabolocom.com} \\
  \And
  Kevin El Haddad\\
  Diabolocom \& ISIA Lab - University of Mons\\
  \texttt{kevin.elhaddad@diabolocom.com } \\
}

\begin{document}

\maketitle

\begin{abstract}
Large language models (LLMs) and high-capacity encoders have advanced zero and few-shot classification, but their inference cost and latency limit practical deployment. We propose training lightweight text classifiers using dynamically generated supervision from an LLM. Our method employs an iterative, agentic loop in which the LLM curates training data, analyzes model successes and failures, and synthesizes targeted examples to address observed errors. This closed-loop generation and evaluation process progressively improves data quality and adapts it to the downstream classifier and task. Across four widely used benchmarks, our approach consistently outperforms standard zero and few-shot baselines. These results indicate that LLMs can serve effectively as data curators, enabling accurate and efficient classification without the operational cost of large-model deployment.
\end{abstract}

\section{Introduction}

Text classification is a longstanding problem in natural language processing~\citep{kowsari2019text, gasparetto2022survey} with wide ranging practical applications. Common use cases include sentiment and stance analysis in social media~\citep{yue2019survey,balahur2013sentiment}, intent detection in conversational agents~\citep{jbene2025intent}, spam and phishing detection in email~\citep{verma2020email, gangavarapu2020applicability}, toxicity and safety moderation~\citep{gunasekara2018review}, customer support ticket routing~\citep{ackerman2023deploying}. Most of the efforts over the past decade were driven by curating large dataset and then training them. However, many real world deployments still face tight constraints on labeling cost, inference latency, and compute budgets. Furthermore, privacy regulations and compliance measures frequently limit access to sensitive user data for model training.

Zero-shot~\citep{yin2019benchmarking, puri2019zero} and few-shot learning~\citep{yu2018diverse, yan2018few} address these constraints by targeting settings with either no labeled examples or only a handful per class. Recent work has largely focused on prompting large language models with label names or descriptions~\citep{wang2023large} and, in the few-shot case~\citep{liu2024knowledge, sun2023text}, a small set of in-context examples. While effective, this strategy depends on expensive LLM inference at test time, leading to high latency and operational cost, along with sensitivity to prompt design and context length. An alternative line of work leverages compact text encoders~\citep{stepanov2025gliclass, boizard2025eurobert} often trained on natural language inference or contrastive objectives to perform classification more efficiently. These models offer lower latency and smaller footprint but typically require well-curated supervision or careful label semantics to perform competitively. Consequently, practitioners face a fundamental trade-off: deploying large models that generalize with minimal data, or utilizing small models that require extensive supervision to achieve competitive accuracy.

To resolve this tension, we build upon recent advances in synthetic data generation~\citep{ye2022zerogen, meng2022generating, wang2021want}, where LLMs act as data curators rather than inference engines. While prior work often relies on one-off static generation, we propose decoupling training from deployment via an iterative, agentic framework. In our approach, an LLM is utilized offline to not just generate but actively refine supervision, enabling the training of a compact classifier that ensures low-latency inference with no test-time LLM dependency. We operationalize this via an iterative, agentic, closed-loop framework. In this cycle, the LLM (i) curates seed training data, (ii) supervises the training of a candidate classifier, (iii) analyzes the classifier’s error modes on the generated set, and (iv) synthesizes targeted hard examples to address the identified deficiencies. By iterating through this generate-evaluate-refine loop, our approach yields increasingly tailored supervision, aligning data quality with the specific inductive biases and needs of the downstream model.

We evaluate our approach on four widely used text-classification benchmarks, utilizing two distinct backbone architectures: SetFit~\citep{tunstall2022efficient}, an efficient few shot training framework for sentence transformers, and EuroBERT~\citep{boizard2025eurobert}, a general-purpose multilingual encoder. Remarkably, in the strictly zero-shot regime (utilizing no human-annotated data), our agentic training procedure outperforms a standard SetFit baseline trained with 8 real labeled examples per class on three of the four datasets. Furthermore, our method consistently surpasses established baselines, including static prompting and zero-shot pre-trained encoders, across both backbones. These findings demonstrate that targeted, LLM-curated supervision can effectively substitute for small labeled datasets, bridging the gap between high-performance generalization and efficient, low-latency inference.

\section{Approach}

\begin{figure}
    \centering
    \includegraphics[width=1\linewidth]{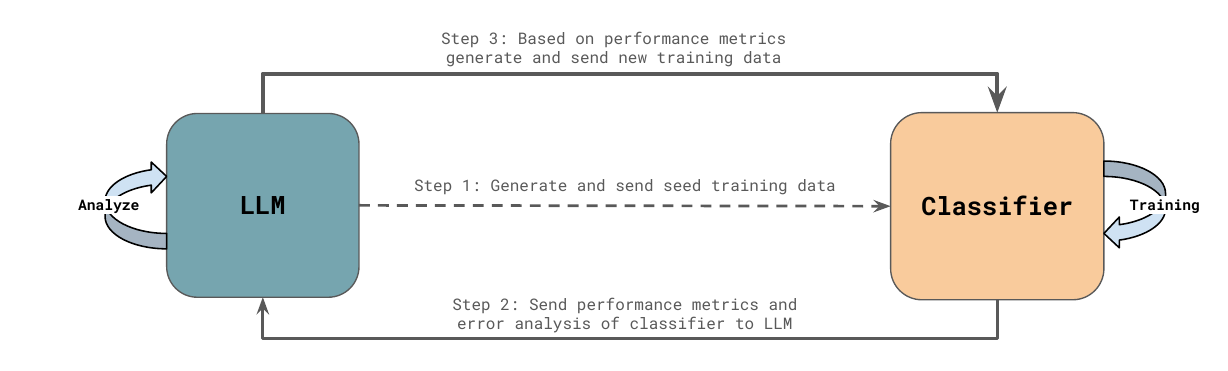}
    \caption{Overview of the proposed agentic training framework. (1) The LLM generates seed training data based on label semantics to train the initial classifier. (2) Performance metrics and specific error modes of the classifier are fed back to the LLM for analysis. (3) The LLM synthesizes targeted examples to address these deficiencies. This cycle repeats for a maximum of K iterations or until convergence, refining the classifier without human supervision.}
    \label{fig:flow}
\end{figure}

We leverage an LLM as a \emph{data curator}: the LLM observes a classifier's errors and synthesizes targeted training examples to correct them. By iterating improving training data, the classifier receives progressively better tailored supervision, yielding higher accuracy without incurring large model cost at test time. This iterative approach is shown in Figure~\ref{fig:flow}

\subsection{Problem Setup}
Let $\mathcal{Y}$ denote the label set and $\mathcal{D}_0$ the initial supervised set. In the \emph{zero-shot} setting, $\mathcal{D}_0=\varnothing$ and only label names or descriptions are provided. While in the \emph{few-shot} setting, $\mathcal{D}_0$ contains $k$ seed examples per label. We assume access to a lightweight classifier $f_\theta$. Our objective is to synthesize a final training set $\mathcal{D}_{\text{final}}$ (constructed iteratively from $\mathcal{D}_0$) that maximizes performance on a held-out test set, subject to a budget on generated examples.

\subsection{Agentic Closed-Loop Framework}
We instantiate a ReAct-style agent~\citep{yao2022react} in which the LLM alternates between planning and invoking tools that we expose to it.

\paragraph{Initialization.}
We provide the agent with: (i) label names; (ii) $\mathcal{D}_0$ (if any); and (iii) instructions to synthesize a \emph{provisional validation set} $\mathcal{D}_{\text{val}}$ that is class-balanced and disjoint from any training data. We also wrap the classifier $f_\theta$ as a callable tool that accepts a training set and a validation set, fits the model, and returns detailed diagnostics (metrics, confusion matrix, per-class error rates, and example-level error traces). Note that the validation set is also generated.

\paragraph{Iterative loop.}
Starting from $\mathcal{D}_{\text{train}} \leftarrow \mathcal{D}_0$, the agent repeats the following generate, evaluate, and refine cycle:
\begin{itemize}
  \item \textbf{Generate:} Given label descriptions, prior examples, and the latest analysis, produce a candidate batch $\mathcal{D}_t$ that is diverse (with constraints on balance, deduplication, and semantic overlap).
  \item \textbf{Train \& Evaluate.} Fit $f_\theta$ on $\mathcal{D}_t$ by calling the classifier tool, and evaluate on $\mathcal{D}_{\text{val}}$ to obtain accuracy, macro-F1, a confusion matrix, and error traces.
  \item \textbf{Analyze.} Summarize success and failure modes (e.g., systematic confusions, lexical cues, negation, domain shift) and propose data needs such as hard negatives, paraphrases, or boundary cases to guide the next synthesis step.
\end{itemize}

The loop terminates when either (i) a maximum number of iterations $T_{\max}$ is reached, or (ii) validation performance plateaus, defined as no improvement greater than a threshold $\epsilon$ for $p$ consecutive iterations. We also enforce a per-class budget on the total number of generated examples. This closed-loop procedure tailors supervision to the classifier and task, improving performance while keeping inference efficient and independent of large-model prompting at test time.

\section{Experiments}

We evaluate the proposed agentic training loop in zero and few-shot text classification. Our aim is to quantify its benefits over standard baselines under limited supervision. To this end, we conduct two sets of experiments. First, we perform an ablation study to isolate the contribution of the agentic loop, comparing the backbone classifiers (\textbf{Baseline}) against static prompting (\textbf{Prompt}) and our proposed method (\textbf{Manager}). Second, we benchmark our approach against established state-of-the-art baselines in few-shot and zero-shot learning (detailed below). Our evaluation utilizes four diverse benchmarks covering fine-grained sentiment analysis, emotion recognition, product reviews, and topic classification:

\begin{itemize}
    \item \textbf{SST-5}~\citep{socher2013recursive}: Five way sentiment classification (very negative, negative, neutral, positive, very positive) on movie review snippets; a standard test of fine-grained sentiment understanding.
    \item \textbf{Emotion}~\citep{saravia-etal-2018-carer}: Six way emotion recognition (anger, disgust, fear, joy, sadness, surprise) from short texts; commonly used for affective analysis with balanced label coverage.
    \item \textbf{CR}~\citep{hu2004mining}: Binary sentiment classification on customer product reviews (positive vs.\ negative), emphasizing opinionated, domain-specific vocabulary and aspect terms.
    \item \textbf{AG News}~\citep{zhang2015character}: Four way topic classification of news headlines and descriptions (World, Sports, Business, Sci/Tech); evaluates general topical discrimination beyond sentiment.
\end{itemize}

In the \emph{few-shot} setting, we uniformly sample $k$ labeled examples per class from the training split and use the \emph{same} sampled set for all baselines and our method to ensure comparability. In the \emph{zero-shot} setting, no labeled examples are provided and only label names are available. We evaluate all models on each dataset’s standard test split. In both zero- and few-shot regimes, we do not use any validation split. Unless otherwise noted, training and loop control rely solely on the available training examples, without access to additional validation data.

We compare our agentic training procedure against the following methods:

\begin{itemize}
    
    \item \textbf{Anchored SetFit (AncSetFit):}~\citep{pauli2023anchoring} Extends SetFit with task/label semantics provided as \emph{anchor statements} (short natural language descriptions). The anchors guide representation learning by injecting semantic priors about the labels.
    
    \item \textbf{ADAPET:}~\citep{tam2021improving} The approach modifies the pattern-exploitation training objective to provide denser supervision during fine-tuning, improving few-shot performance.
    
    \item \textbf{GliClass:}~\citep{stepanov2025gliclass} A modern, off-the-shelf zero-shot text classifier trained using Gliner like architecture \footnote{\url{https://huggingface.co/knowledgator/gliclass-modern-base-v3.0}}; evaluated in its recommended zero-shot configuration.

\end{itemize}

\subsection{Implementation Details}

We implement the agent using the Hugging Face \texttt{smolagents} framework~\citep{smolagents}, employing its \texttt{CodeAgent} to orchestrate the generate--evaluate--refine loop. We rely on GPT-5 for all our LLM based tasks. The downstream classifier is SetFit with the 110M-parameter sentence encoder \texttt{sentence-transformers/all-mpnet-base-v2}. We use the reference SetFit training pipeline and the library’s default hyperparameters without tuning, as zero-/few-shot regimes often do not provide a labeled validation split. Unless otherwise noted, the same configuration is applied across all datasets. For all experiments, we run for three different seeds and report the average performance. AncSetFit and ADAPET-Base results are from \citet{pauli2023anchoring}. For reproducibility, we have released code, prompts, and experiment scripts here\footnote{\url{https://github.com/Diabolocom-Research/data-generation}}.

\begin{table*}[t]
\centering
\caption{Zero and few-shot accuracy (\%) on four text-classification benchmarks using SetFit and Encoder backbones. 
$k$ denotes the number of labeled examples per class. 
The ``Manager'' columns denote our proposed agentic loop. 
Best result per row is in \textbf{bold}.}
\label{tab:ablation-results}
\setlength{\tabcolsep}{4pt}
\begin{tabular}{ll ccc c ccc}
\toprule
 &  & \multicolumn{3}{c}{\textbf{SetFit Backbone}} & & \multicolumn{3}{c}{\textbf{Encoder Backbone}} \\
\cmidrule{3-5} \cmidrule{7-9}
Dataset & $k$ & Baseline & +Prompt & +Manager & & Baseline & +Prompt & +Manager \\
\midrule
\multirow{4}{*}{SST-5}
`& 0 & --   & 43.2 & \textbf{47.0} & & --   & 22.5 & \textbf{24.2} \\
& 2 & 33.4 & 38.8 & \textbf{46.38} & & 22.9   & 25.9 & \textbf{26.4} \\
& 4 & 38.4 & \textbf{46.2} & 46.1 & & 18.5   & 24.4 & \textbf{27.4} \\
& 8 & 42.8 & \textbf{47.7} & 46.2 & & 24.6   & 24.3 & \textbf{30.0} \\
\addlinespace
\hline
\addlinespace
\multirow{4}{*}{Emotion}
& 0 & --   & 40.3 & \textbf{50.9} & & --   & 23.7 & \textbf{35.12} \\
& 2 & 30.0 & 39.1 & \textbf{56.5} & & 17.2   & \textbf{29.3} & 25.73 \\
& 4 & 49.0 & 54.3 & \textbf{56.0} & & 24.0   & \textbf{27.3} & 21.22 \\
& 8 & 45.0 & 56.1 & \textbf{61.6} & & 20.9   & \textbf{33.2} & 26.42 \\
\addlinespace
\hline
\addlinespace
\multirow{4}{*}{CR}
& 0 & --   & 87.2 & \textbf{87.8} & & --  & 50.0 & \textbf{64.9} \\
& 2 & 75.5 & \textbf{89.5} & 89.0 & & 54.7   & \textbf{58.7} & 52.5 \\
& 4 & 86.8 & 89.1 & \textbf{90.1} & & 58.5   & 66.7 & \textbf{67.7} \\
& 8 & 89.6 & \textbf{89.5} & 89.4 & & 58.2   & \textbf{69.1} & 68.2 \\
\addlinespace
\hline
\addlinespace
\multirow{4}{*}{AG News}
& 0 & --   & 73.0 & \textbf{82.6} & & --     & 71.8 & \textbf{76.7} \\
& 2 & 65.3 & 69.4 & \textbf{83.2} & & 60.0   & 71.1 & \textbf{79.0} \\
& 4 & 76.5 & 82.6 & \textbf{85.0} & & 75.5   & 78.7 & \textbf{79.7} \\
& 8 & 82.8 & 81.0 & \textbf{84.36} & & 76.2  & 77.2 & \textbf{80.1} \\
\bottomrule
\end{tabular}
\end{table*}

\subsection{Results}
We present results related to impact of agentic supervision in Table~\ref{tab:ablation-results}, comparing the backbone classifier (\textit{Baseline}) against static prompting strategies (\textit{Baseline+Prompt}) and our proposed agentic loop (\textit{Baseline+Manager}). Following previous works, we report accuracy on the standard test split for each dataset.

We observe that our agentic approach consistently outperforms both the naive baseline and static prompting across most settings. This advantage is most pronounced in the zero-shot regime, where the iterative generation of targeted synthetic data provides a significant initialization benefit. Furthermore, while static prompting improves over the baseline, our method consistently achieves higher accuracy, validating the efficacy of the error-driven feedback loop over one-off generation. Regarding the underlying architecture, we observe that the SetFit backbone consistently outperforms the general-purpose EuroBERT encoder. This aligns with standard intuition that the SetFit's contrastive fine-tuning mechanism is specifically optimized for few-shot sentence similarity, whereas standard encoders require more data to stabilize. Given this distinct performance gap, the remainder of our analysis focuses primarily on the SetFit backbone to evaluate our method within the highest-performing regime. In Table~\ref{tab:main-results} we compare our method against established zero-shot and few-shot baselines. For the sake of brevity, we only present results related to the SetFit backbone, as it showed stronger performance in our first experiment.

\begin{table}[t]
\centering
\begin{threeparttable}
\caption{Zero and few-shot accuracy (\%) on four text-classification benchmarks. 
$k$ denotes the number of labeled examples per class. 
``Ours'' is the proposed agentic loop. 
Best result for each ($\text{dataset},k$) row is in \textbf{bold}. AncSetFit and ADAPET-Base results are from \citet{pauli2023anchoring}.}
\label{tab:main-results}
\setlength{\tabcolsep}{8pt} % Increased spacing slightly since columns were removed
\begin{tabular}{llrrrr}
\toprule
Dataset & $k$ & AncSetFit & ADAPET-Base & GliClass 3.0 & \textbf{Ours} \\
 &  & (110M) & (117M) & (151M) & (110M) \\
\midrule
\multirow{4}{*}{SST-5}
& 0 & --   & --   & 43.8 & \textbf{47.0} \\
& 2 & 40.5 & 32.9 & --   & \textbf{46.38} \\
& 4 & 43.4 & 35.5 & --   & \textbf{46.2} \\
& 8 & 45.2 & 39.6 & --   & \textbf{47.7} \\
\addlinespace
\hline
\addlinespace
\multirow{4}{*}{Emotion}
& 0 & --   & --   & \textbf{51.9} & 50.9 \\
& 2 & 49.1 & 34.5 & --   & \textbf{56.5} \\
& 4 & 53.2 & 45.0 & --   & \textbf{56.0} \\
& 8 & 50.9 & 53.9 & --   & \textbf{61.6} \\
\addlinespace
\hline
\addlinespace
\multirow{4}{*}{CR}
& 0 & --   & --   & \textbf{90.2} & 87.8 \\
& 2 & 86.6 & 76.8 & --   & \textbf{89.01} \\
& 4 & 89.8 & 77.0 & --   & \textbf{90.1} \\
& 8 & \textbf{90.7} & 78.6 & --   & 89.4 \\
\addlinespace
\hline
\addlinespace
\multirow{4}{*}{AG News}
& 0 & --   & --   & 68.8 & \textbf{82.6} \\
& 2 & 76.0 & 64.5 & --   & \textbf{83.2} \\
& 4 & 81.6 & 73.1 & --   & \textbf{85.0} \\
& 8 & 84.2 & 77.9 & --   & \textbf{84.4} \\
\bottomrule
\end{tabular}
\end{threeparttable}
\end{table}

Zero-shot Performance: In the absence of human-labeled data (k=0), our agentic training procedure outperforms the larger GLiClass~3.0 model (151M parameters) on three of the four datasets, despite utilizing a significantly smaller 110M parameter encoder (roughly 73\% of the size). The gains are particularly notable on \textsc{AG~News}, suggesting that our iterative curation effectively captures complex label semantics that off-the-shelf zero-shot models may miss. However, GLiClass retains an edge on \textsc{Emotion}, likely because its pre-training aligns well with standardized sentiment tasks. Note that we exclude SetFit, AncSetFit, and ADAPET from this comparison as they do not natively support zero-shot inference.

Few-shot Performance: In few-shot settings \(k \in {2,4,8}\), our approach demonstrates robust generalization. With as few as k=2 examples per class, we observe consistent improvements over the standard SetFit baseline and outperform ADAPET-Base across all evaluated settings. Notably, we observe a trend of diminishing returns as k increases from 2 to 8 on most datasets. We hypothesize that for broad, common categories (as in \textsc{SST-5} or \textsc{CR}), our targeted synthetic supervision saturates the model's capacity early. \textsc{AG~News} remains the exception, where increased supervision continues to yield gains, likely due to the task's specific, information-dense label semantics which benefit from additional guidance.

\section{Conclusion}

We introduced a closed-loop \emph{generate--evaluate--refine} framework for text classification in which a ReAct-style LLM agent curates supervision. It synthesizes training data, trains a lightweight classifier, diagnoses errors, and targets subsequent generation to address observed failure modes. By decoupling training from deployment, the approach enables accurate, efficient inference without relying on large models at test time. Across four benchmarks in zero and few-shot regimes, our method consistently outperforms or matches strong baselines, highlighting the effectiveness of LLMs as data curators rather than test-time predictors.

In the future, we aim to extend the framework to structured prediction tasks (e.g., NER, relation extraction), and explore alternative downstream learners. We also plan to explore richer analysis tools (e.g., calibration- and robustness-aware diagnostics), incorporate human-in-the-loop checks, and study quality trade-offs under different LLM backbones and generation budgets.

\bibliographystyle{plainnat}
\bibliography{references}

\end{document}